\documentclass{article}

\usepackage[preprint]{corl_2026} % Uncomment for pre-prints (e.g., arxiv); This is like ``final'', but will remove the CORL footnote.

\makeatletter
\if@preprinttype
  \renewcommand{\@notice}{}
\fi
\makeatother

\usepackage{algorithm}
\usepackage[noend]{algpseudocode}
\usepackage{amsmath}
\usepackage{graphicx}
\usepackage{tabularx}
\usepackage{booktabs}
\usepackage{xcolor}
\usepackage[caption=false,font=footnotesize]{subfig}

\newcommand{\para}[1]{\noindent \emph{#1}}

\title{The Bitter Lesson Comes to Robotics}

\newcommand{\inspectauthorblock}{%
  \parbox[t]{\dimexpr\textwidth-2\tabcolsep\relax}{%
    \centering
    \normalfont\fontsize{10}{12}\selectfont
    \setlength{\parskip}{0pt}
    \mbox{Christopher Leet\textsuperscript{1,2}}\quad
    \mbox{Achyutha Menon\textsuperscript{1}}\quad
    \mbox{Sravanthi Machcha\textsuperscript{1}}\quad
    \mbox{Sven Koenig\textsuperscript{3}}\\
    \mbox{Tzu Kit Chan\textsuperscript{1}}\quad
    \mbox{Jay Chooi\textsuperscript{1}}
    \par\vspace{5pt}
    \fontsize{9}{11}\selectfont
    \mbox{\textsuperscript{1}Robocurve}\quad
    \mbox{\textsuperscript{2}University of Southern California}\quad
    \mbox{\textsuperscript{3}University of California, Irvine}
    \par
  }%
}
\author{\inspectauthorblock}

\begin{document}
\maketitle

% Explicit PDF metadata, independent of the formatted author block.
\hypersetup{
  pdftitle={The Bitter Lesson Comes to Robotics},
  % pdfauthor={}, % For the initial submit
  pdfauthor={Christopher Leet, Achyutha Menon, Sravanthi Machcha, Sven Koenig, Tzu Kit Chan, Jay Chooi},
  pdfsubject={We systematically evaluate the ability of multimodal LMs directly controlling robotic hardware to perform manipulation tasks.}
}

%===============================================================================

\begin{abstract}
General purpose multimodal language models (LMs) are increasingly capable of directly controlling robotic arms (e.g., by commanding end effector poses) to manipulate their environment. Tracking the development of these capabilities is important for understanding their potential benefits, such as improving industrial automation, and risks, such as enabling malicious users to produce dangerous chemical or biological agents. To this end, we investigate the ability of frontier multimodal LMs to directly control physical robotic arms. We evaluate nine models at multiple reasoning-effort and service-tier settings on six manipulation tasks taken from the RoboDojo benchmark. To our knowledge, this is the first systematic comparison of a broad range of frontier LMs across a common suite of manipulation tasks on physical hardware. We find that current frontier models can directly control real robotic arms to perform a variety of manipulation tasks but are not yet precise or efficient enough to complete them within a practical action budget. However, these capabilities are improving rapidly: performance increases on our benchmark between July and October 2026 (inclusive) correspond to a performance doubling time of 53.9 days, while the best recorded execution rate increased by a factor of 4.4 and the lowest recorded execution cost fell by a factor of 6.2. These trends raise the possibility that the Bitter Lesson is coming to robotics; the capabilities of embodied general purpose LMs may one day be comparable to those of more specialized models.
\end{abstract}

% Two or three meaningful keywords should be added here
\keywords{ Physical Language Model Benchmarking, Embodied AI.}

%===============================================================================
\section{Introduction}

General purpose multimodal language models (LMs) are increasingly capable of manipulating physical environments by directly controlling robotic arms. For example, Zhang et al.~\cite{Zhang2026UnexpectedRobotPolicy} demonstrate that GPT-6 Astra can grasp a variety of objects by specifying successive end-effector targets, that is, positions, orientations, and openings for the robot's grippers. Tracking the growth of these capabilities is important for understanding their potential benefits and risks.

To date, however, evaluations of LM manipulation capabilities on physical hardware have either focused on a small number of models or compared multiple models on only one or two manipulation tasks. Systematic comparisons of multiple models on multiple tasks run in the real world remain scarce. Simulation, however, may not fully capture physical effects such as calibration drift, camera parallax, and arm sag, which can substantially affect model performance. Moreover, evaluating a range of models on the same tasks and hardware is essential for tracking capability improvements.

We address this gap by evaluating nine frontier LMs' performance, inference costs, and execution rate on six manipulation tasks from the RoboDojo benchmark using a real I2RT YAM bimanual robot~\cite{i2rt_yam}. We capture partial progress by dividing each task into a series of subtasks and scoring models based on the subtasks that they completed. Subtasks provide a more granular measure of performance than task success and allow us to identify specific manipulations that models struggle to complete. We measure inference cost and execution rate in terms of average cost per subtask and average subtasks completed per hour. These metrics allow us to draw conclusions about model efficiency from partially completed tasks.  

We find that frontier models can directly control robotic arms to perform a variety of physical manipulation tasks, but are currently not precise or efficient enough to complete them reliably within a practical action budget. We also find, however, that performance is improving rapidly. Performance increases on our benchmark from July through October 2026 correspond to a performance doubling time of 53.9 days. Moreover, the best recorded subtask completion rate increased by $4.4 \times$ and the lowest recorded inference costs per subtask decreased by $6.2 \times$ over that period. This suggests that, in the near future, the performance, cost and execution time achieved by general purpose multimodal LMs may be comparable to those of specialized robotic systems: \emph{The Bitter Lesson}~\cite{Sutton2019BitterLesson} may be coming to robotics.

\section{Related Work}

\para{Evaluating the Manipulation Capabilities of Embodied General-Purpose LMs.} Several recent studies evaluate whether frontier multimodal LMs can directly control robotic hardware. Zhang et al.~\cite{Zhang2026UnexpectedRobotPolicy} study whether GPT-6 Astra, GPT-5.5 Sol, and DeepSeek-Flash can control bimanual robots via end-effector pose commands. They evaluate these models on all 42 tasks in the RoboDojo~\cite{Chen2026RoboDojo} benchmark in simulation and find that Astra outperforms all 40 published robot policy entries. They also test Astra on real robotic arms but halt the experiments after 33 recorded trials because Astra repeatedly commanded unsafe actions, leading to hardware damage.

Jia et al.~\cite{Jia2026AgentAsPolicy} evaluate GPT-6 Astra on eight task configurations of five manipulation tasks, including assembly, targeted throwing, and towel folding, using physical I2RT YAM~\cite{i2rt_yam} robotic arms. They allow Astra both to command joint-space or Cartesian end-effector targets directly and to write and execute Python programs that compute and submit motion commands. They find that Astra achieves success rates of at least 80\% in seven of eight task configurations, although execution is expensive and slow (tens of minutes and dollars). They additionally compare Astra with five other models on a two-pair assembly task and find that it achieves a 100\% success rate (along with GPT-5.6 Sol and Opus 5) and achieves the shortest mean completion time (only counting successful trials). On a block construction task, Astra achieves a higher success rate than GPT-5.6 Sol.

Berman et al.~\cite{Berman2026ClaudePlaysRobotics} evaluate 12 models' performance in manipulation (e.g., moving objects), locomotion (e.g., balancing a Unitree Go2 quadruped~\cite{Unitree2023Go2}), and classical control (e.g., pendulum balancing) in simulation. Their manipulation experiments use 40 tasks from the LIBERO~\cite{Liu2023LIBERO} benchmark and a seven-DoF Franka Panda~\cite{Haddadin2022FrankaEmikaRobot} arm. They also perform exploratory experiments using a real Unitree Go2 quadruped, primarily centered on navigation. They study both direct control (where their models issue low-level actions) and indirect control (where their models code, train, and direct policies). They find that direct control is improving but remains unreliable compared to indirect control.

While these studies demonstrate that general-purpose LMs can control robots, systematic comparisons of their ability to perform manipulation tasks on physical hardware remain limited. Such evaluations are important because physical robots often have imperfections such as sag, calibration drift, and positioning scatter, which may not be accurately captured in simulation. In this paper, we evaluate the ability of frontier LMs to directly control physical robotic arms. To our knowledge, this is the first systematic comparison of a broad range of frontier LMs across a common suite of manipulation tasks on physical hardware. We evaluate nine models in fifteen configurations on six tasks from RoboDojo. We use the same embodiment and experimental interfaces for each evaluation, allowing us to compare performance, execution time, and cost under shared experimental conditions.

\para{Tracking the Capabilities of LMs.} Our analysis of progress over time is inspired by prior work tracking capability improvements. Kwa et al.~\cite{Kwa2025LongSoftwareTasks} track how the length of the software tasks that LM agents can complete at a given success rate increases over time. They use the time required for human experts to complete each task to define its length. We track how manipulation capabilities are increasing over time. We measure progress as performance on a set of physical tasks.

Performance alone does not determine an agent's practical usefulness. Kapoor et al.~\cite{Kapoor2024AgentsThatMatter} argue that agent evaluations should also measure cost since this often affects which systems are preferable. We apply this perspective to robotic control by evaluating performance, cost, and wall-clock time, allowing us to characterize the tradeoffs between these three dimensions.
\section{Methodology}
\begin{figure*}[t]
    \centering

    % Horizontal gap between neighboring images.
    \newcommand{\robodojogap}{0.3mm}%

    % Each image occupies one third of the current image-grid width,
    % after accounting for the two gaps in each row.
    \newcommand{\robodojopanel}[2]{%
        \begin{minipage}[t]{%
            \dimexpr(\linewidth-\robodojogap*2)/3\relax}
            \centering
            \vspace{0pt}%
            \includegraphics[width=\linewidth]{#1}%
            \par\vspace{3pt}%
            {\small\bfseries #2\par}%
        \end{minipage}%
    }%

    % LEFT: six starting scenes in a three-column, two-row grid.
    \begin{minipage}[t]{0.55\textwidth}
        \centering
        \vspace{0pt}%

        % First row.
        \makebox[\linewidth][c]{%
            \robodojopanel
                {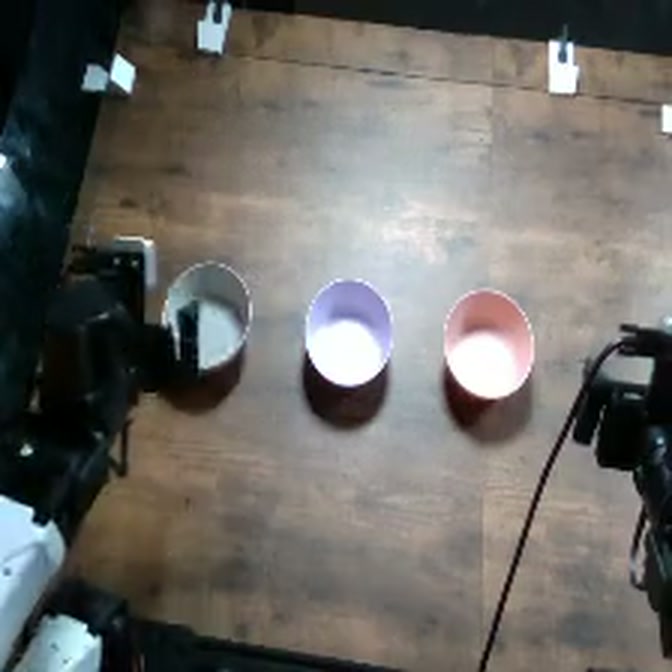}
                {Stack Bowls}%
            \hspace{\robodojogap}%
            \robodojopanel
                {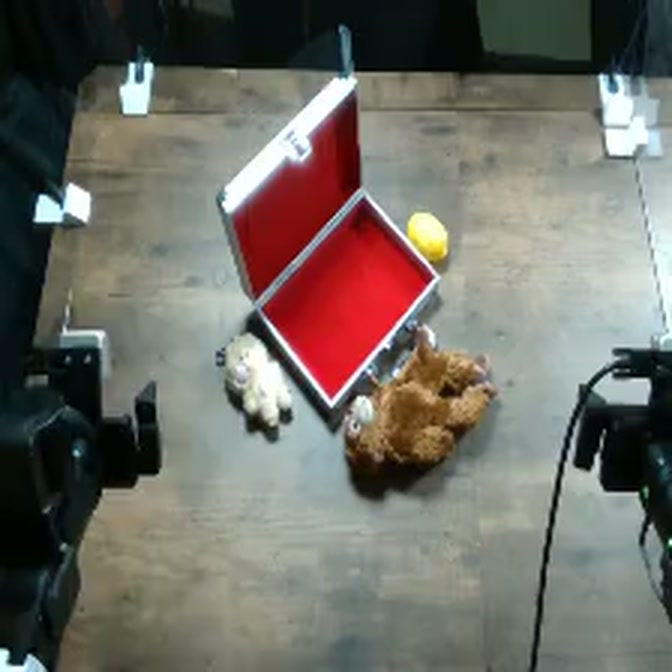}
                {Store in Safe}%
            \hspace{\robodojogap}%
            \robodojopanel
                {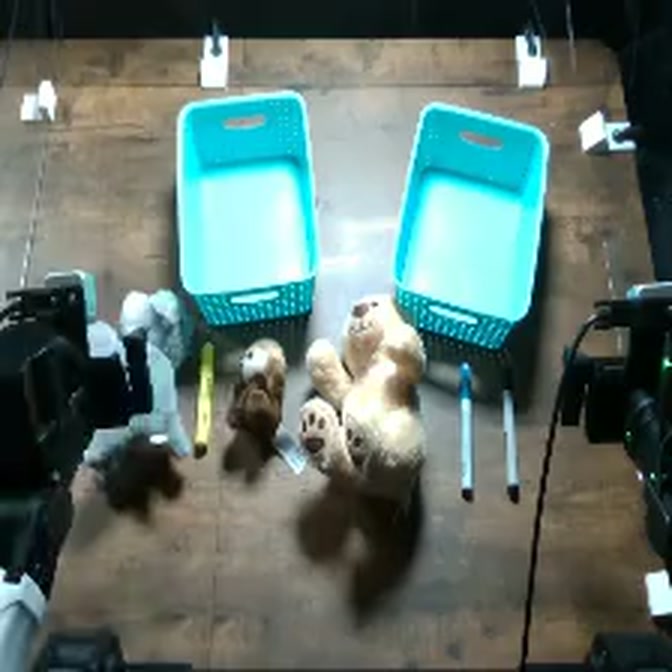}
                {Sort Objects}%
        }%

        \par\vspace{2mm}%

        % Second row.
        \makebox[\linewidth][c]{%
            \robodojopanel
                {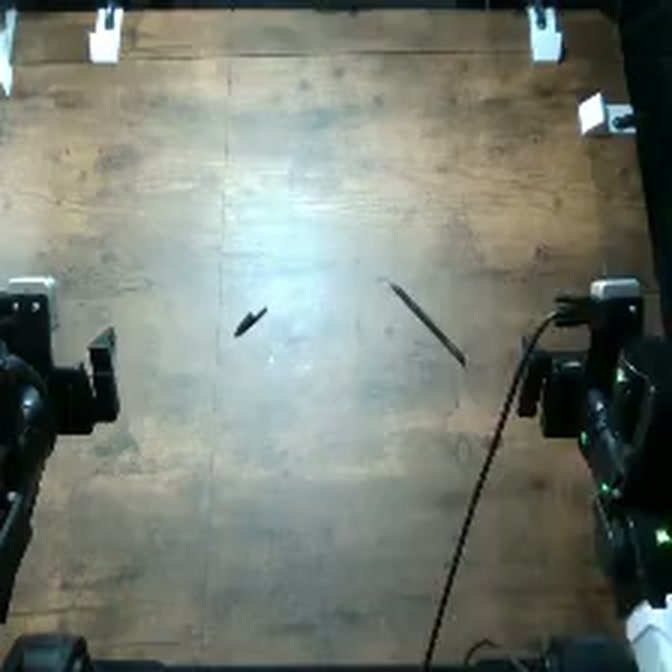}
                {Cap Pen}%
            \hspace{\robodojogap}%
            \robodojopanel
                {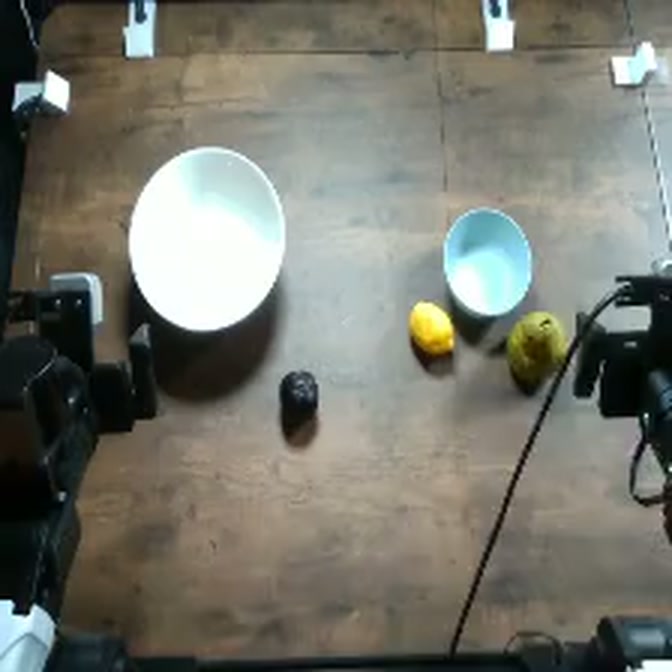}
                {Pack and Pour\\Fruit}%
            \hspace{\robodojogap}%
            \robodojopanel
                {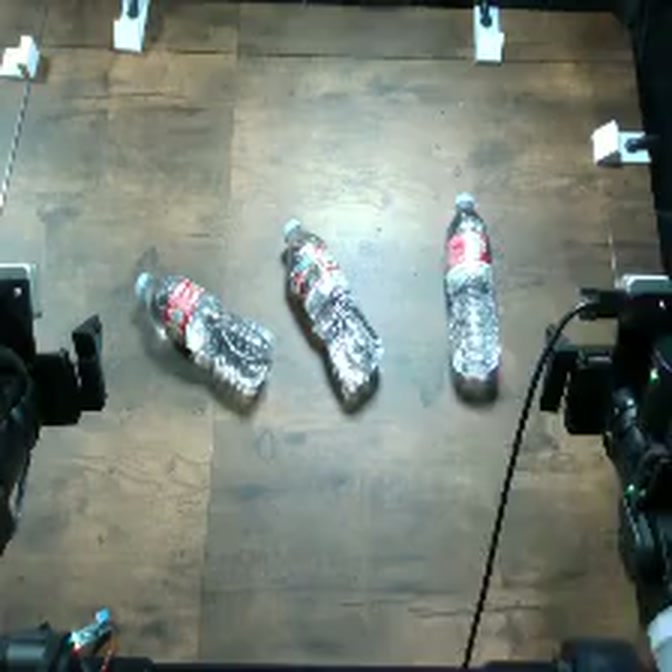}
                {Stand Up Bottles}%
        }%
    \end{minipage}%
    \hfill%
    % RIGHT: task instructions, ordered to match the image grid.
    \begin{minipage}[t]{0.43\textwidth}
        \vspace{0pt}%
        \footnotesize
        \setlength{\tabcolsep}{3pt}%
        \renewcommand{\arraystretch}{1.1}%

        \begin{tabularx}{\linewidth}{%
            @{}
            >{\raggedright\arraybackslash\bfseries}p{0.29\linewidth}
            >{\raggedright\arraybackslash}X
            @{}
        }
            \toprule
            Task & \textbf{Instruction} \\
            \midrule

            Stack Bowls &
            Stack the bowls on the table. \\
            \addlinespace[3pt]

            Store in Safe &
            Place all the objects into the safe, then close
            the safe. \\
            \addlinespace[3pt]

            Sort Objects &
            Classify the two types of objects on the table
            and place them into the two baskets. \\
            \addlinespace[3pt]

            Cap Pen &
            Pick up the pen and the pen cap, then put the
            cap on the pen. \\
            \addlinespace[3pt]

            Pack and Pour Fruit &
            Place all the fruits into the blue bowl, then
            pour the fruits from the blue bowl into the
            large white bowl. \\
            \addlinespace[3pt]

            Stand Up Bottles &
            Stand all three bottles upright. \\

            \bottomrule
        \end{tabularx}
    \end{minipage}

    \caption{Starting scenes (left) and task instructions (right) for the six benchmark tasks.}
    \vspace{-4mm}
    \label{fig:robodojo-tier1-start-scenes}
\end{figure*}

We run our evaluations on six tasks taken from the RoboDojo~\cite{Chen2026RoboDojo} benchmark: (a) stack bowls, (b) store in safe, (c) sort objects, (d) cap pen, (e) pack and pour fruit, and (f) stand up bottles. Each task's initial scene and instruction are shown in Figure~\ref{fig:robodojo-tier1-start-scenes}.

Each evaluation is conducted on a bimanual I2RT YAM~\cite{i2rt_yam} robot. This robotic platform provides policies with two types of observations: (a) proprioceptive feedback from its arms and grippers and (b) three camera feeds. Two of these cameras are mounted on the robot's wrists, and the third is positioned above the robot to capture a top down view of the robot's workspace. The images in Figure~\ref{fig:robodojo-tier1-start-scenes} are frames captured by the overhead camera during our experiments.

We evaluate nine frontier multimodal LMs. Three of these models, Astra, Opus 5.5, and GPT-6.1 Sol, were evaluated in four, three, and two configurations, respectively, for a total of fifteen (model, configuration) pairs. We use the suffixes (M), (H), and (XH) to denote medium, high, and extra-high reasoning effort respectively. The suffixes (H+UF) and (H+PR) denote high reasoning effort with ultrafast service and priority service respectively. We evaluate four OpenAI models: GPT-5.6 Sol (H), GPT-6 Sol (H), GPT-6 Astra (M, H, XH, H+UF), and GPT-6.1 Sol (H, H+UF); two Anthropic models: Opus 5 (M) and Opus 5.5 (M, H, XH); and three Google models: Gemini 3.6 Flash (H), Gemini 3.8 Flash (H), and Gemini 4 Argon (H+PR). Each (model, configuration) pair was evaluated on each task over 20 trials.

The Inspect Robots~\cite{Leet2026InspectRobots} framework was used to give the models control of the robotic platform. At the start of each trial, the framework provides the model with the task instruction, a description of how to operate the robot, and observations of the initial scene. The framework then operates a closed loop. At the start of a loop, the model selects an action. The framework executes the action and appends the observations collected during execution to the model's context. This process repeats until the model terminates the trial, exhausts its budget of 40 actions or causes a hardware fault.

Our framework provides models with three types of actions: \texttt{move\_to}, \texttt{done}, and \texttt{give\_up}. Each action is provided to the models as a tool. The \texttt{move\_to} action provides Cartesian end-effector control over the robot's two arms. It allows models to submit an $(x, y, z, \mathrm{yaw}, \mathrm{pitch}, \mathrm{roll}, \mathrm{gripper})$ 7-tuple for each arm. This tuple specifies the position and orientation of its gripper in the arm's base coordinate frame as well as the gripper's aperture. Gripper apertures are specified on the normalized range $[0,1]$. The framework uses inverse kinematics to convert each target end effector pose into a joint configuration, a tuple which specifies the angle of each joint in each arm. The \texttt{done} action allows the model to signal task completion. The \texttt{give\_up} action allows the model to terminate a trial without completing the task. \texttt{give\_up} allows the model to indicate if an unrecoverable error has made the task impossible to complete and refuse unsafe tasks. The action requires the model to explain why it is terminating the trial.

% ~\cite{Craig2005}

Each joint configuration is reviewed by a set of guards before it is realized. Each guard can approve, modify, or reject the configuration. Our evaluations use two guards. The first clips configurations to the arms' operating limits. For example, if a configuration requires a gripper to open wider than its hardware allows, the guard clips the configuration to the gripper's maximum aperture. The second restricts the maximum distance that an arm can move in one action. This guard prevents sudden large movements that could damage equipment or injure nearby people.

We evaluate models on three metrics: task performance, inference cost, and wall clock time. To measure partial progress, we divide each task into a sequence of subtasks. Each subtask is associated with a score between 0\% and 100\% which represents cumulative progress towards task completion. The subtasks themselves are cumulative, that is, subtasks are designed so that completing any given subtask requires completing all lower scoring subtasks. For example, we divide the task ``store in safe'' into four subtasks:  \emph{1 object in safe}, \emph{2 objects in safe},  \emph{3 objects in safe}, and \emph{3 objects in safe, lid closed} worth 25\%, 50\%, 75\% and 100\% respectively. Figure~\ref{fig:robodojo-survival-curves-models} shows each task's subtasks and their associated task scores. Subtasks provide us with a more granular measure of performance than tracking binary task success or failure and allow us to identify the specific stages of each task where models struggle. Each trial is graded by an instance of GPT-6 Astra (H) based on the footage taken during the trial by the robot's three cameras. A trial receives the highest score associated with a subtask that was completed at some point during the trial, even if the model undoes its progress and the subtask is incomplete by the end of the trial.

\section{Evaluating Task Performance}
\vspace{-2mm}

\begin{figure*}[t]
    \centering \includegraphics[width=\textwidth]{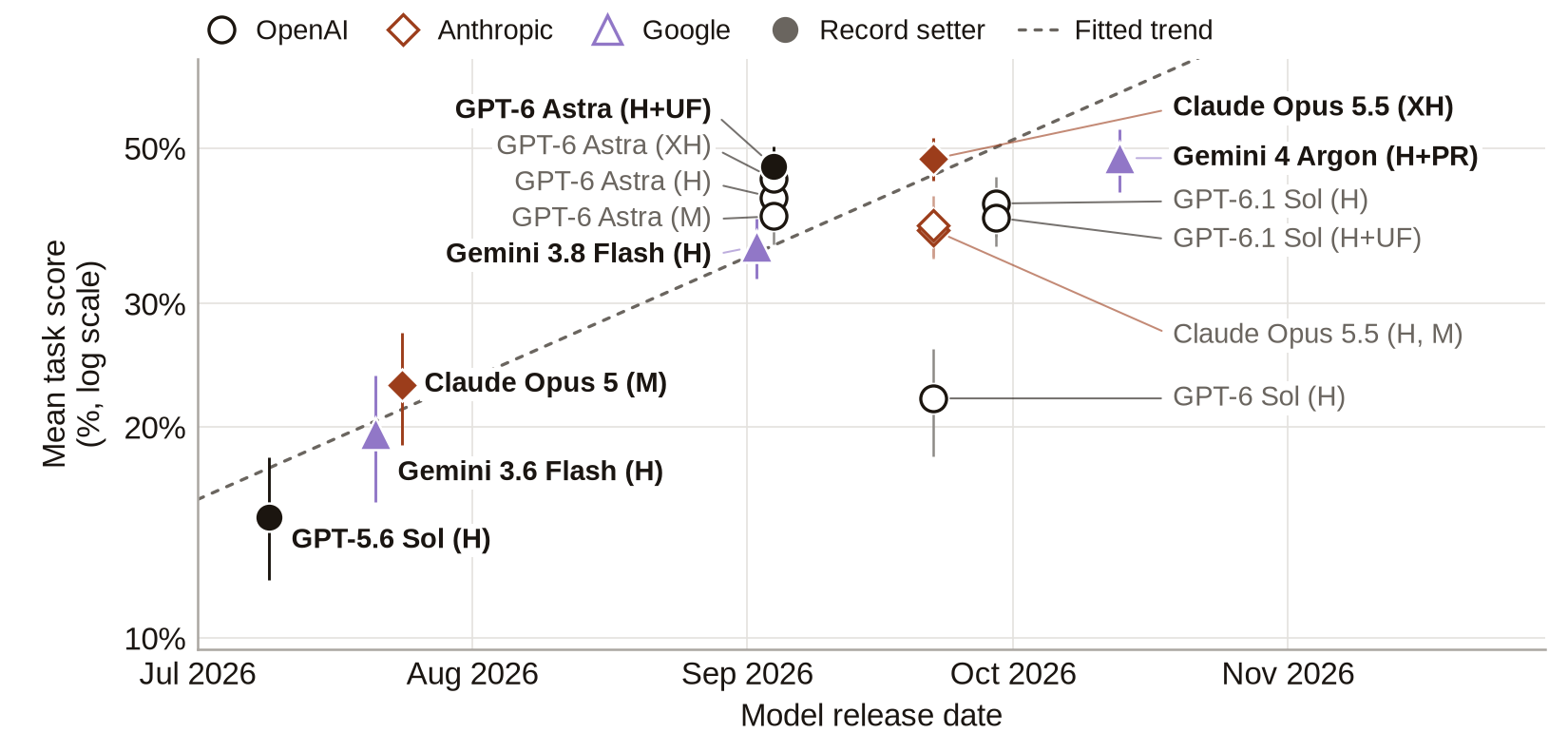}
    \vspace{-2mm}
    \caption{The mean task score achieved by each (model, configuration) pair in our evaluation against model release dates. The error bars indicate each task score's 95\% bootstrap confidence interval. Record setting pairs, pairs which outscored every other previously and concurrently released pair, are plotted with filled markers and have bolded labels.}
    \vspace{-3mm}
    \label{fig:metr-plot}
\end{figure*}

Figure~\ref{fig:metr-plot} plots each (model, configuration) pair's mean task score against its release date. Mean task score is computed by averaging the score awarded to a (model, configuration) pair across all $6\, \text{tasks} \times 20\, \text{trials} = 120$ trials. Recently released models typically achieve significantly higher scores, indicating widespread improvements in manipulation capability. Record setting pairs, pairs which outscored every other previously and concurrently released pair, are plotted with filled markers and have bolded labels. Following the approach of METR~\cite{Kwa2025LongSoftwareTasks}, we estimate the rate at which performance is increasing by fitting a least-squares regression of the logarithm of the mean task score achieved by each record-setting (model, configuration) pair against their model's release date. The fitted trend corresponds to a performance doubling time of 53.9 days over the time interval July 2026 - October 2026.

We estimate uncertainty using a task-stratified percentile bootstrap. We generate 20,000 bootstrap replicates by resampling 20 trial scores with replacement from each of the six tasks for each (model, configuration) pair. We compute the mean scores achieved by each (model, configuration) pair in each replicate, identify the record setting pairs, and refit the regressions. The 2.5th and 97.5th percentile doubling times of these regressions give us a 95\% bootstrap confidence interval of 36.1--59.5 days. The relatively wide confidence interval reflects uncertainty about whether Gemini 4 Argon (H+PR) is a new record setter. Gemini 4 Argon outscores Opus 5.5 (XH) by just 0.13 points. In 45\% of bootstrap replicates, Gemini 4 Argon (H+PR) is a record setter, leading to doubling time estimates centered around 53 days. In the other 55\%, every record setting model is released by September 22, producing estimates centered around 42 days. 

We assess the sensitivity of our doubling time estimate to each record setting (model, configuration) pair by omitting each pair and then refitting the regression. The doubling time estimates that we obtain range from 44.3 to 61.1 days.

\begin{figure*}[t]
    \centering \includegraphics[width=\textwidth]{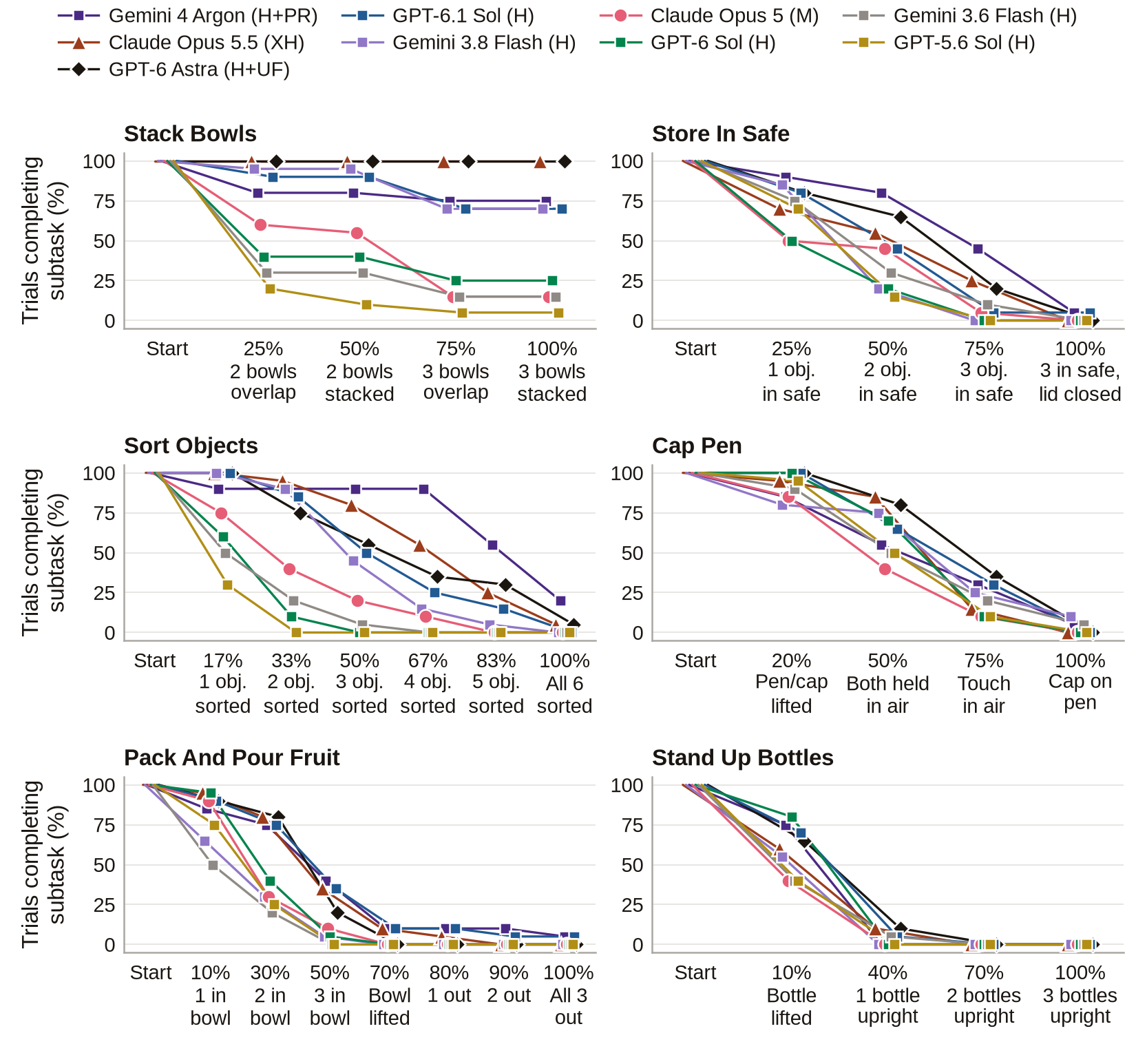}
    \caption{The survival curves generated by the configuration with the highest mean task score for each model. Each survival curve shows the percentage of trials that (possibly temporarily) accomplished each subtask associated with a task. The x-axes show each task's subtasks and their associated scores.}
    \label{fig:robodojo-survival-curves-models}
    \vspace{-5mm}
\end{figure*}

Despite this rapid improvement, task completion was relatively rare. While the strongest models achieved mean scores of 45\% -- 48\%, these scores reflect consistent partial progress rather than intermittent end-to-end success. Figure~\ref{fig:robodojo-survival-curves-models} shows the survival curves associated with each task. Each panel plots the fraction of trials of each model that reached each subtask, beginning with the initial task state on the left, and ending with full task completion on the right. These curves show how far each model typically progresses through the task during a trial. We only profile the top performing configuration of Opus 5.5 (Opus 5.5 (XH)), Astra (GPT-6 Astra (H+UF)) and GPT-6.1 Sol (GPT-6.1 Sol (H)). (GPT-6 Astra (H+UF) outperformed GPT-6 Astra (XH), and GPT-6.1 Sol (H) outperformed GPT-6.1 Sol (H+UF), by 1.9 percentage points each.)

The survival curves show that the models fall into two categories. GPT-6 Astra, GPT-6.1 Sol, Opus 5.5, Gemini 3.8 Flash, and Gemini 4 Argon typically make substantial progress on their tasks while Opus 5, GPT-5.6 Sol, GPT-6 Sol, and Gemini 3.6 Flash perform less well. These categories are visible in Figure~\ref{fig:metr-plot}. Analyzing these models' trajectories suggests that the two sets of models are largely differentiated by their efficiency. Models rarely get stuck: they usually know what to do next and how to recover when an action fails. Every model typically makes steady progress until it succeeds, exhausts its action budget, or makes an unrecoverable error, such as knocking an object off the table. Consequently, our task scores tend to reflect how much each model can accomplish with a limited action budget, rather than what it can achieve given unlimited time. (The task Stand Up Bottles, discussed below, is an exception.)

Consequently, increasing the action budget would almost certainly improve completion rates for all models. However, we consider a limited action budget important for evaluating practical performance. As discussed in Section~\ref{sec:time-and-cost}, current trials already take several minutes; trajectories requiring hundreds of actions would take substantially longer. A policy that completes a task after twenty minutes is very different, in practical terms, from a policy that completes it efficiently. Our action budget lets us measure whether a policy can complete a task efficiently enough to be useful.

An inability to grasp objects efficiently is a major contributor to the poor performance of the weaker models. For example, GPT-5.6 Sol and GPT-6 Sol frequently fail to position their grippers in the right location for a successful grasp. These models close their grippers without grasping an object 64\% and 59\% of the time, compared to 32--39\% for Astra, Opus 5.5, and Gemini 3.8 Flash. (GPT-6.1 Sol falls between these groups with a 46\% failure rate.) Opus 5 is good at finding the right position for its gripper (its grasp attempts fail 33\% of the time, comparable to the stronger models) but uses a lot of actions in the process. It takes an average of 6.7 actions to position its gripper, rapidly consuming its action budget. It also frequently overestimates the width of its targets. 19\% of its grasp attempts specify an aperture of 0.1 or more (on a normalized $[0,1]$ scale), which often leaves the jaws of its gripper too far apart to grasp the object.

% Notably, GPT-5.6 Sol and GPT-6 Sol often position their gripper too high. GPT-5.6 Sol's and GPT-6 Sol's median grasp pose heights are 3.6--3.9 cm, compared to 2.3--2.8 cm for Astra, Opus 5.5, and GPT-6.1 Sol. 22\% of their grasp attempts occur above 6~cm; 72\% of these grasps are empty. Even when a grasp attempt connects, gripping an object too high often leads to the grip failing. For example, gripping above the center line of a fruit or bottle can cause the gripper's jaws to slide off its curved surface. Their transcripts suggest that these errors may reflect a failure to compensate for physical effects such as camera parallax, which can distort perceived distances as the wrist cameras approach an object.

\begin{figure*}[t]
    \centering \includegraphics[width=\textwidth]{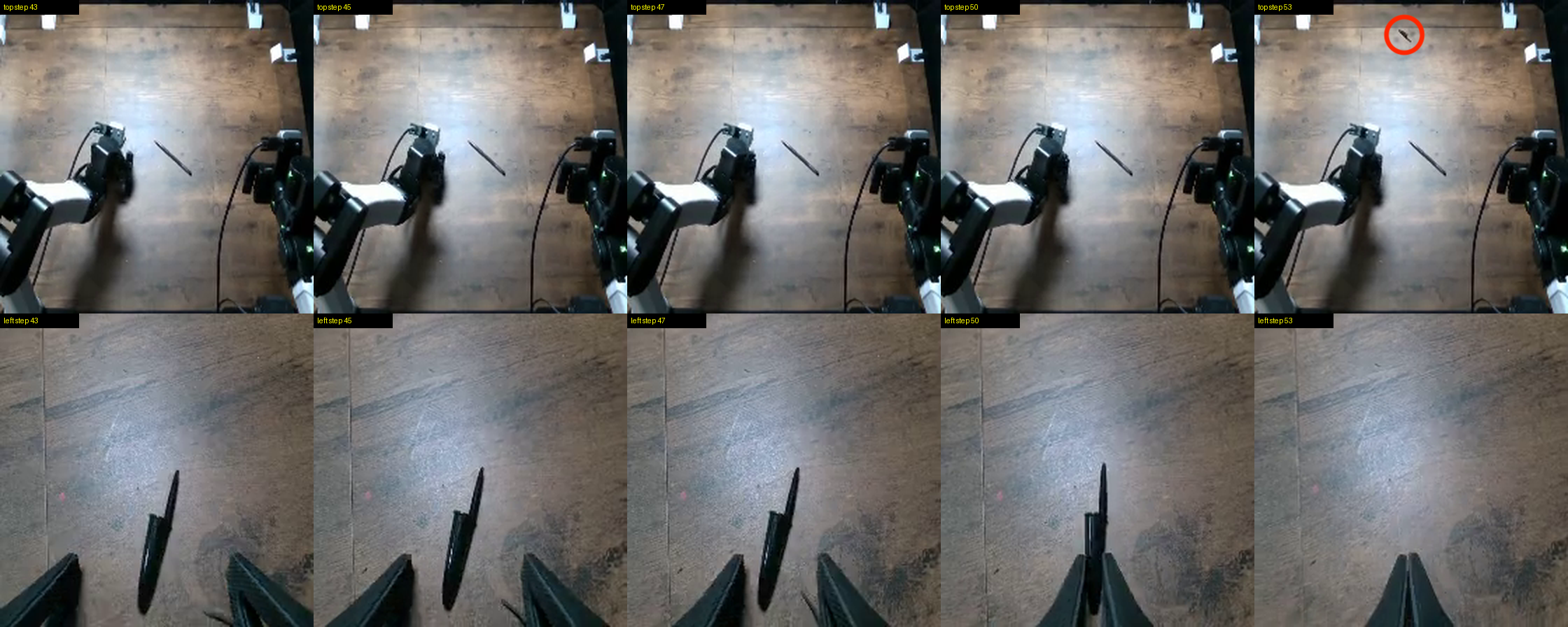}
    \caption{An example unrecoverable error: Opus 5.5 accidentally launches a pen lid out of reach. The top and bottom panels show footage from the platform's top and left wrist cameras, respectively.}
    \vspace{-1mm}
    \label{fig:poor-opus}
\end{figure*}

Every model was tripped up by the imperfections inherent to physical systems, such as arm sag and camera parallax. The stronger models, however, were able to recognize, characterize, and compensate for these effects more effectively than the weaker models. The task Stack Bowls is an illustrative example. All five stronger models completed the task in at least 70\% of their trials while the four weaker models completed it in at most 25\% of their trials. Every model recognized that the bowls were wider than the grippers' maximum opening and that therefore pinching a side wall was the only way to lift a bowl. Only the five more successful models, however, executed this grasp reliably. Since the walls of the bowls were thin, the gripper produced a reading of 0.02--0.05 (on a normalized $[0,1]$ scale) after a successful pinch, indistinguishable from an empty pinch. The only reliable way to verify a grasp was therefore to try lifting the bowl. Stronger models consistently performed this check, while weaker models often misinterpreted the gripper reading and wasted actions regrasping securely held bowls.

Because of their ability to diagnose problems and execute operations efficiently, when a stronger model receives a low score, it is usually because an unrecoverable error has made further progress impossible. Examples include knocking an object off the table, bumping the safe shut before all objects are inserted, or launching the pen lid out of reach by squeezing its curved surface. Figure~\ref{fig:poor-opus} depicts one such error (a pen lid launch). The top and bottom rows show successive frames from the platform's top camera and left wrist camera, respectively.

Stand Up Bottles is an exception to this pattern. Our setup made this task unusually difficult, and it is therefore unclear whether increasing the action budget would allow the models to complete it reliably. The bottles were positioned relatively far away from the robot's base. As a result, grasping the bottles without first repositioning them brought the arm's shoulder and elbow joints close to full extension. This sharply limited the set of poses that the model could use to grasp the bottles, making it difficult to find a grasp which allowed the bottles to be rotated upright in a single action. Moreover, end effector pitch was limited to $\pm 0.6$ radians, further complicating rotation. Consequently, only 6 out of 180 trials across our nine (model, configuration) pairs shown in Figure~\ref{fig:robodojo-survival-curves-models} resulted in one bottle being stood upright, and no trial resulted in two. These findings suggest that even frontier models struggle to find and reliably execute solutions to difficult manipulation tasks.

\begin{figure*}[t]
    \centering \includegraphics[width=\textwidth]{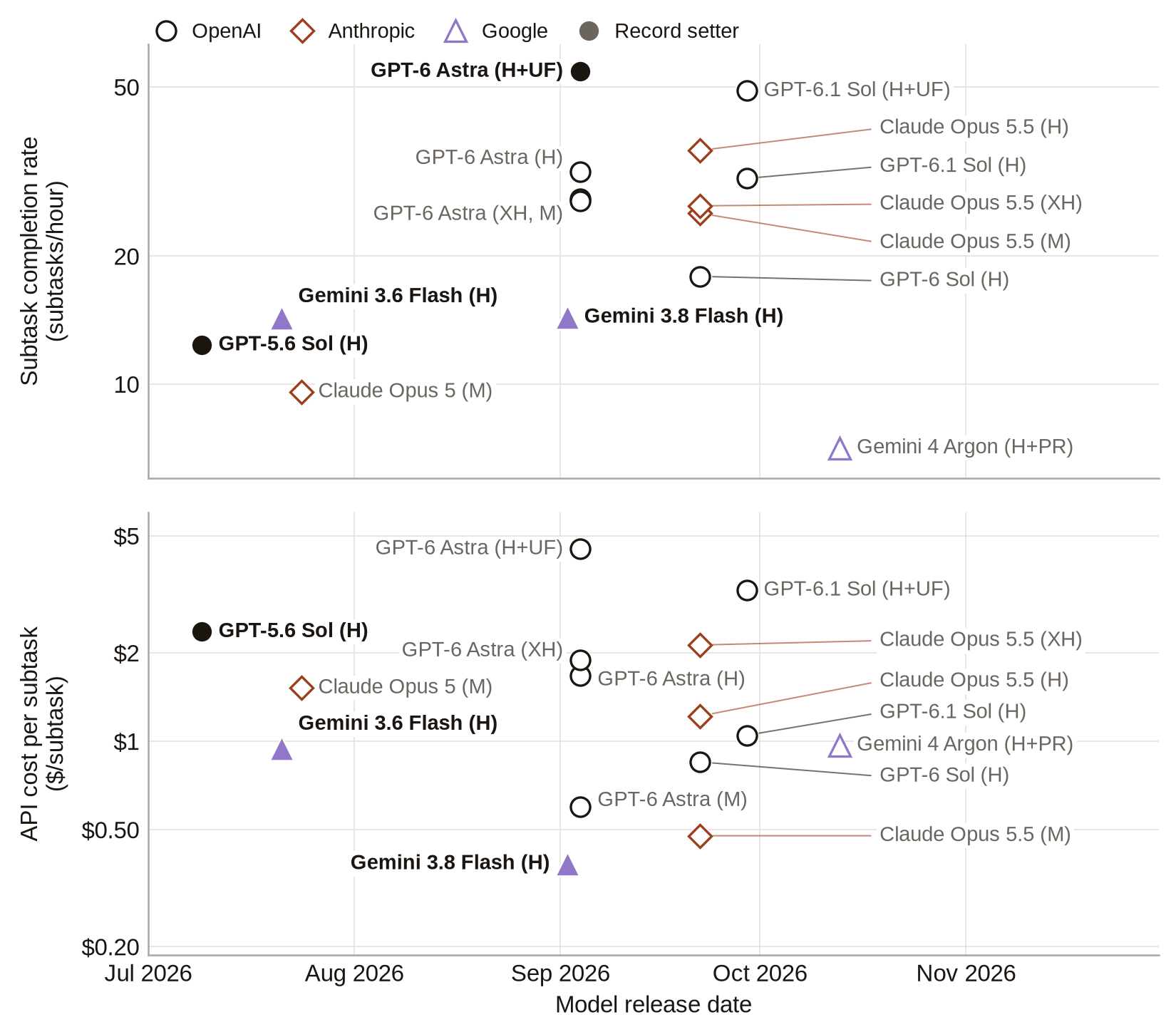}
    \vspace{-7mm}
    \caption{The subtask completion rate (top), measured in subtasks per hour, and estimated API cost per subtask (bottom), measured in dollars per subtask, achieved by each (model, configuration) pair in our evaluations against model release date. Record setting pairs, pairs that achieved a higher subtask completion rate (top) or lower cost per subtask (bottom) than every other previously or concurrently released pair, are plotted with filled markers and have bolded labels.}
    \vspace{-3mm}
    \label{fig:cost-and-time}
\end{figure*}

These experiments demonstrate that frontier models can directly control robotic arms to perform a variety of physical manipulation tasks, but are currently not precise or efficient enough to complete them reliably within a practical action budget. Moreover, these models often fail to identify strategies which can accommodate the limitations of their embodiment. However, these experiments also show that performance is improving rapidly; future generations of models may overcome these limitations.

% On every other task, even the best models only make it to 50\% completion less than 80\% of the time. Each of these tasks, however, had failure modes that made completion impossible, such as bumping the safe door shut, knocking an object off the table, or launching a pen lid across the room by squeezing it on its curved face. These failure modes largely account for the low scores achieved by the top model. 
\section{Evaluating Subtask Completion Rate and Inference Cost}
\label{sec:time-and-cost}

Figure~\ref{fig:cost-and-time} plots the subtask completion rate (top), measured in subtasks per hour, and the estimated API cost per subtask, measured in dollars per subtask, achieved by each (model, configuration) against the model's release date. We measure a (model, configuration) pair's subtask completion rate by dividing the total number of subtasks that it completed during all 120 trials by the total wall clock time that it spent executing those trials (i.e., action execution time and inference time). We compute its API cost per subtask by dividing the total API cost of executing its trials by the number of subtasks that it completed. API cost is computed from the number of uncached input, cache read, cache write, and output tokens consumed during each run, using the rates given by each provider. These metrics were designed to account for the cost and time associated with unsuccessful attempts at completing subtasks.

Considered in isolation, both execution time and inference costs have been decreasing rapidly. The best mean number of subtasks completed per hour recorded increased from 12.4 (achieved by GPT-5.6 Sol (H), released on July 9th) to 54.4 (achieved by GPT-6 Astra (H+UF), released on Sept 4th): a $4.4\times$ increase over 57 days. Moreover, the best mean API cost per subtask recorded decreased from \$2.36 (achieved by GPT-5.6 Sol (H), released on July 9th) to \$0.38 (achieved by Gemini 3.8 Flash (H), released Sept 2nd): a $6.2\times$ decrease over a similar time period (55 days). We estimate the rate at which models have been improving on each metric by fitting a log-linear regression to the datapoints associated with each record-setting (model, configuration) pair. The fitted trends correspond to a subtask completion rate doubling time of 45.4 days and an API cost per subtask halving time of 23.2 days. Since, however, there were only four record setting pairs for the completion rate and three for cost, we view these estimates as exploratory.

\begin{figure*}[t]
    \centering \includegraphics[width=\textwidth]{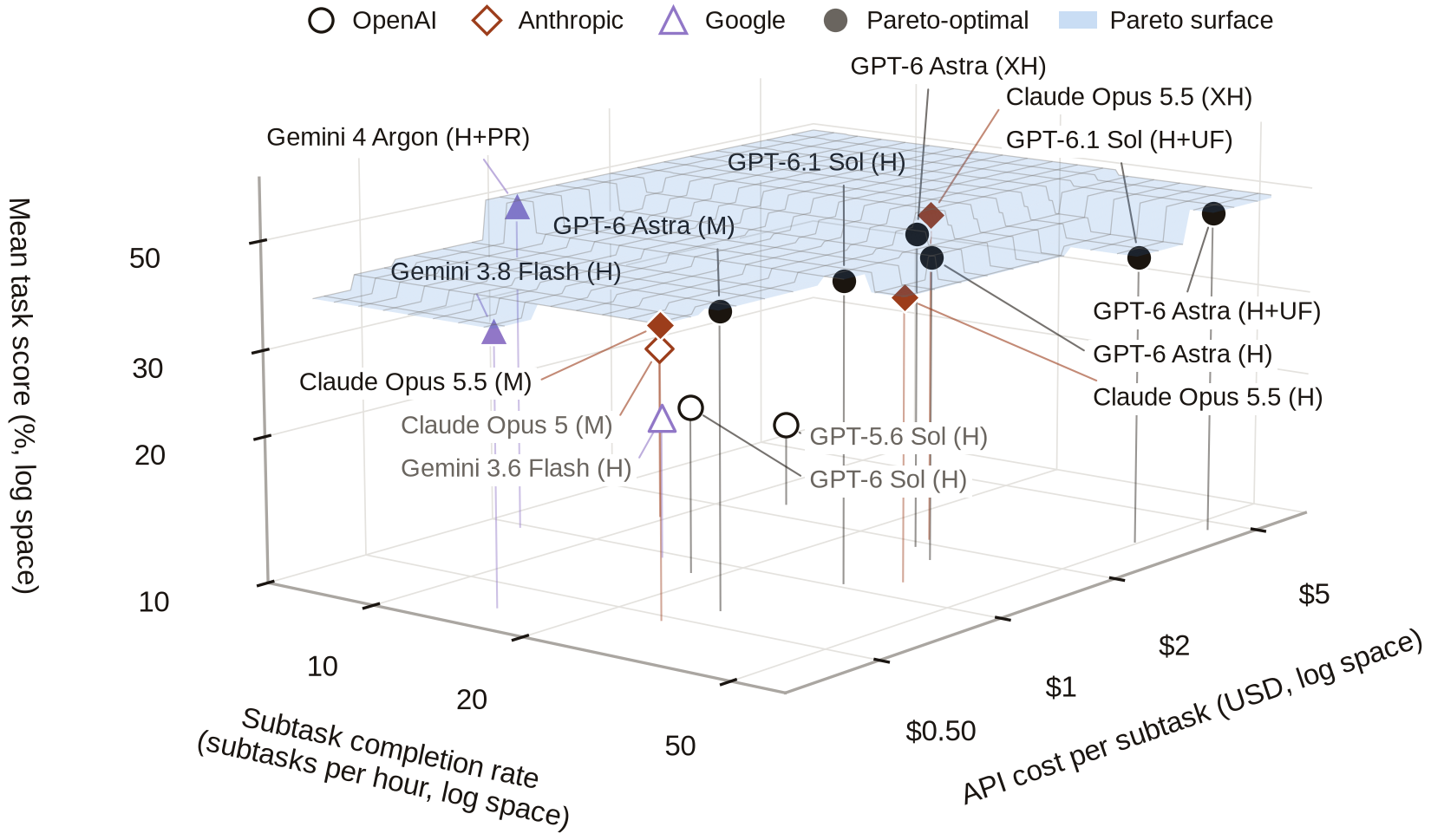}
    \caption{The mean task score, subtask completion rate (measured in subtasks per hour) and estimated API cost per subtask (measured in dollars per subtask) achieved by each (model, configuration) pair in our evaluations. The blue surface depicts the Pareto frontier defined by the pairs whose performance, subtask completion rate and cost are not dominated by any other pair. Models on the Pareto frontier have filled markers; models off it have hollow markers. All three axes use a logarithmic scale.}
    \vspace{-4mm}
    \label{fig:robodojo-pareto-frontier}
\end{figure*}

% Inference cost and wall clock time were usually more dependent on action efficiency than raw inference costs and latency. 

Considering execution time and inference costs in isolation can be misleading. Fast execution time is often associated with high inference costs and vice versa. We show how these two metrics trade off against each other and performance in Figure~\ref{fig:robodojo-pareto-frontier}. Figure~\ref{fig:robodojo-pareto-frontier} plots the mean task score achieved by our models against their subtask completion rate and the API cost per subtask. The Pareto surface, depicted in blue, is estimated using the envelope of the models whose (performance, cost, completion rate) metrics are not dominated by those of any other model. Figure~\ref{fig:robodojo-pareto-frontier} plots nondominated models  with filled markers and dominated models with hollow markers.

Today, no single model (or even group of models) dominates: there are clear tradeoffs between performance, cost, and time. GPT-6 Astra (H+UF) completes a subtask every 1.1 minutes on average, 11\% faster than the second fastest model, but costs \$4.51 per subtask to run, 38\% more costly than the second most expensive model. Conversely, Gemini 3.8 Flash (H) costs just \$0.38 per subtask to run, but takes 4.18 minutes to complete a subtask on average. Other notable models include:

\emph{Gemini 4 Argon (H+PR).} This model  achieves the highest performance on our benchmark (48.4\%, 0.13 points above Opus 5.5 (XH)) at a moderate cost: just \$0.97 per subtask. However, it is also the slowest model by a wide margin, completing just 7.1 subtasks per hour. By constrast, Claude Opus 5 (M), the second slowest model, completes 9.6 subtasks per hour. Argon's high reasoning latency can leave the robot's arms in positions that likely require high joint torques to counteract gravity for extended periods of time, occasionally causing an arm to overheat and terminating the trial early.

\emph{Opus 5.5 (H) and GPT-6 Astra (H).} These models are notably efficient: they are the third and fourth fastest models overall (executing 35.5 and 31.5 subtasks per hour), mid-priced (they cost \$1.22 and \$1.67 per subtask to run) and offer reasonable performance (38.2\% and 42.5\%).

\section{Conclusion}

Our evaluations show that frontier multimodal LMs can directly control robotic arms to perform a variety of manipulation tasks. Currently state-of-the-art LMs, however, are not precise or efficient enough to execute these tasks within a realistic time budget. Moreover, their inference costs are still relatively high and subtask completion rates are modest. Despite this, model capabilities are improving rapidly on all three metrics. A log-linear fit to the scores achieved by the record setting (model, configuration) pairs released between July and October 2026 corresponds to a performance doubling time of 53.9 days. Over the same time period, the best recorded subtask completion rate increased by a factor of 4.4, while the lowest estimated API cost per subtask fell by a factor of 6.2. These trends raise the possibility that the \emph{Bitter Lesson}~\cite{Sutton2019BitterLesson} is coming to robotics. As general-purpose multimodal LMs improve, they may become competitive with specialized robotic policies in performance, execution time, and cost.

%===============================================================================

% no \bibliographystyle is required, since the corl style is automatically used.
\bibliography{example}  % .bib

\end{document}